\def\year{2022}\relax
\documentclass[letterpaper]{article}
\usepackage{aaai22} 
\usepackage{times} 
\usepackage{helvet} 
\usepackage{courier} 
\usepackage[hyphens]{url} 
\usepackage{graphicx}
\usepackage{natbib}  
\usepackage{caption} 
\DeclareCaptionStyle{ruled}{labelfont=normalfont,labelsep=colon,strut=off} 
\usepackage{algorithm}
\usepackage{algorithmic}

\usepackage{newfloat}
\usepackage{listings}
\floatstyle{ruled}
\newfloat{listing}{tb}{lst}{}
\floatname{listing}{Listing}

\nocopyright

\usepackage{xspace}
\usepackage{graphicx}
\usepackage[utf8]{inputenc}
\usepackage[T1]{fontenc}
\usepackage{booktabs}
\usepackage{amsfonts}
\usepackage[fleqn]{amsmath}
\usepackage{nicefrac}
\usepackage{microtype}
\usepackage[inline]{enumitem}
\usepackage{tikz}
\usepackage{multirow}
\usepackage{eucal}
\usepackage{comment}
\usepackage{bm}
\usepackage{subcaption}
\usepackage{mathtools}
\usepackage{xcolor,colortbl}
\usepackage{url}

\def\Re{\mathbb{R}}

\def\Nat{{\rm I\kern\pIR N}}

\newcommand{\EE}[1]{\exptE\left[#1\right]}

\def\PBE{\overline{\text{PBE}}}

\newcommand{\defeq}{\overset{\text{\tiny def}}{=}}

\def\MSVEm{\overline{\text{VE}}}
\def\MSVE{$\overline{\text{VE}}$\xspace}

\def\A{{\mathcal{A}}}

\def\R{{\mathcal{R}}}
\def\S{{\mathcal{S}}}

\newcommand{\States}{\S}

\def\vec0{{\bf{0}}}
\def\vecb{{\bf{b}}}

\def\vecv{{\bf{v}}}
\def\vecw{{\bf{w}}}
\def\vecx{{\bf{x}}}

\def\vecz{{\bf{z}}}

\def\w{\vecw}

\newcommand{\Amat}{\mathbf{A}}

\newcommand{\beq}{\begin{equation}}
\newcommand{\eeq}{\end{equation}}
\newcommand{\beqa}{\begin{eqnarray}}
\newcommand{\eeqa}{\end{eqnarray}}
\newcommand{\beqan}{\begin{eqnarray*}}
\newcommand{\eeqan}{\end{eqnarray*}}
\newcommand{\ben}{\begin{eqnarray*}}
\newcommand{\een}{\end{eqnarray*}}

\def\tr{^\top\!}

\def\vecw{{\bf{\bf w}}}
\def\vecz{{\bf{\bf z}}}
\def\vecx{{\bf{\bf x}}}

\def\vecv{{\bf{\bf v}}}
\renewcommand{\EE}[2]{\mathbb{E}_{#1\!\!}\left[#2\right]}

\newcommand{\CEE}[3]{\EE{#1}{{#2}~\middle\vert~{#3}}}
\renewcommand{\CEE}[3]{\EE{#1}{{#2}\mid{#3}}}

\def\CEpi#1#2{\CEE{\pi}{#1}{#2}}

\newcommand{\vhat}{\hat{v}}

\def\la{($\lambda$)\xspace}
\def\l{$\lambda$\xspace}
\def\a{$\alpha$\xspace}
\def\ze{($\zeta$)\xspace}
\def\lab{($\lambda,\beta$)\xspace}

\def\hp{Rooms\xspace}
\def\hv{High Variance Rooms\xspace}
\def\MSVEm{\overline{\text{VE}}}
\def\AVEm{\text{A}\overline{\text{VE}}}
\def\MSVE{$\overline{\text{VE}}$\xspace}
\def\AVE{$\text{A}\overline{\text{VE}}$\xspace}

\title{An Empirical Comparison of Off-policy Prediction Learning Algorithms \\ in the Four Rooms Environment}
\author {
    Sina Ghiassian,\textsuperscript{\rm 1}
    Richard S. Sutton \textsuperscript{\rm 2}
}
\affiliations {
    \textsuperscript{\rm 1} University of Alberta\\
    \textsuperscript{\rm 2} University of Alberta and DeepMind \\
    ghiassia@ualberta.ca,
    rsutton@ualberta.ca
}

\begin{document}

\maketitle

\begin{abstract}
Many off-policy prediction learning algorithms have been proposed in the past decade, but it remains unclear which algorithms learn faster than others. In this paper, we empirically compare 11 off-policy prediction learning algorithms with linear function approximation on two small tasks: the \hp task, and the \hv task. The tasks are designed such that learning fast in them is challenging. In the \hp task, the product of importance sampling ratios can be as large as $2^{14}$ and can sometimes be two. To control the high variance caused by the product of the importance sampling ratios, step size should be set small, which in turn slows down learning. The \hv task is more extreme in that the product of the ratios can become as large as $2^{14}\times 25$. This paper builds upon the empirical study of off-policy prediction learning algorithms by Ghiassian and Sutton (2021). We consider the same set of algorithms as theirs and employ the same experimental methodology. The algorithms considered are: Off-policy TD\la, five Gradient-TD algorithms, two Emphatic-TD algorithms, Tree Backup\la, Vtrace\la, and ABTD\ze. We found that the algorithms' performance is highly affected by the variance induced by the importance sampling ratios. The data shows that Tree Backup\la, Vtrace\la, and ABTD\ze are not affected by the high variance as much as other algorithms but they restrict the effective bootstrapping parameter in a way that is too limiting for tasks where high variance is not present. We observed that Emphatic TD\la tends to have lower asymptotic error than other algorithms, but might learn more slowly in some cases. We suggest algorithms for practitioners based on their problem of interest, and suggest approaches that can be applied to specific algorithms that might result in substantially improved algorithms.
\end{abstract}


\section{Off-policy Prediction Learning}
\label{sct:OffPolicyLearning}

To learn off-policy is to learn about a target policy while behaving using a different behavior policy.
In prediction learning, the policies are given and fixed.
In this paper, we conduct a comparative empirical study of 11 off-policy prediction learning algorithms on two tasks.

Off-policy learning is interesting for many reasons from learning options (Sutton, Precup, \& Singh, 1999), to auxiliary tasks (Jaderberg et al., 2016), and learning from historical data (Thomas, 2015).
One interesting use-case of off-policy learning is learning about many different policies in parallel (Sutton et al., 2011), which we consider in this paper.

In previous work, many algorithms have been developed for off-policy prediction learning.
Off-policy TD\la uses importance sampling ratios to correct for the differences between the target and behavior policies but it is not guaranteed to converge (Precup, Sutton, \& Singh, 2000).
Later, Gradient and Emphatic-TD algorithms were proposed to guarantee convergence under off-policy training with linear function approximation (Sutton et al., 2009; Sutton, Mahmood, \& White, 2016).
These convergent algorithms were later developed further to learn faster. Proximal GTD2\la, TDRC\la, and Emphatic TD\lab are examples of such algorithms (Mahadevan et al., 2014; Ghiassian et al., 2020; Hallak et al., 2016).
Another interesting group of algorithms exclusively focuses on learning fast, rather than convergence. An example of such algorithm is Vtrace\la (Espeholt et al., 2018).

Off-policy learning has been essential to many of the recent successes of Deep Reinforcement Learning (Deep RL).
The DQN architecture (Mnih et al., 2015) and its successors such as Double DQN (van Hasselt, Guezm, \& Silver, 2016) and Rainbow (Hessel et al., 2018) rely on off-policy learning.
The core of many of these architectures is Q-learning (Watkins, 1989), the first algorithm developed for off-policy control.
Recent research used some modern off-policy algorithms such as Vtrace and Emphatic-TD within Deep RL architectures (Espeholt et al., 2018; Jiang et al., 2021), but it remains unclear which of the many off-policy learning algorithms developed to date empirically outperforms others.

Unfortunately, due to the computational burden, it is impossible to conduct a large comparative study in a complex environment such as the Arcade Learning Environment (ALE).
The original DQN agent (Mnih et al., 2015) was trained for one run with a single parameter setting.
A detailed comparative study, on the other hand, needs at least 30 runs; typically includes a dozen algorithms, each of which have their own parameters.
For example, to compare 10 algorithms on the ALE, each with 100 parameter settings (combinations of step-size parameter, bootstrapping parameter, etc.), for 30 runs, we need 30,000 times more compute than what was used to train the DQN agent on an Atari game.
One might think that given the increase in available compute since 2015, such a study might be feasible.
Moore's law states that the available compute approximately doubles every two years.
That means compared to 2015, eight times more compute is at hand today.
Taking this into account, we still need 30,000/8=3750 times more compute than what was used to train one DQN agent.
This is simply not feasible now, or in the foreseeable future.

Let us now examine the possibility of conducting a comparative study in a state-of-the-art domain, similar to Atari, but smaller.
MinAtar (Young \& Tian, 2019) simplifies the ALE environment considerably, but presents many of the same challenges.
To evaluate the possibility of conducting a comparative study in MinAtar, we compared the training time of two agents.
One agent used the original DQN architecture (Mnih et al., 2015), and another used the much smaller Neural Network (NN) architecture of Young and Tian (2019) used for training in MinAtar.
Both agents were trained for 30,000 frames on an Intel Xeon Gold 6148, 2.4 GHz CPU core.
On average, each MinAtar training frame took 0.003 of a second (0.003s) and each ALE training frame took 0.043s.
To speed up training, we repeated the same procedure on an NVidia V100SXM2 (16GB memory) GPU.
Each MinAtar training frame took 0.0023s and each ALE training frame took 0.0032s.
The GPU did a good job speeding up the process that used a large NN (in the ALE), but did not provide much of a benefit on the smaller NN used in MinAtar.
This means, assumign we have enough GPUs to train on, using MinAtar and ALE will not be that different.
Given this data, detailed comparative studies in an environment such as MinAtar are still far out of reach.

The most meaningful empirical comparisons have in fact been in small domains.
Geist and Scherrer (2014) was the first such study that compared all off-policy prediction learning algorithms to date.
Their results were complemented by Dann, Neumann, and Peters (2014) with one extra algorithm and a few new problems.
Both studies included quadratic and linear computation algorithms.
White and White (2016) followed with a study on prediction learning algorithms but narrowed down the space of algorithms to the ones with linear computation, which in turn allowed them to go into greater detail in terms of sensitivity to parameters.
The study by Ghiassian and Sutton (2021) also focused on linear computation algorithms.
They introduced a small task, called the \emph{Collision task} and applied 11 algorithms to it.
They explored the parameter space in detail and studied four extra algorithms.
Taking into account the final performance, learning speed, and sensitivity to various parameters, Ghiassian and Sutton (2021) grouped the 11 algorithms into three tiers.
We will discuss their grouping in more detail later in the paper.

This paper conducts a comparative study of off-policy prediction learning algorithms with a focus on the variance issue in off-policy learning.
The structure of the paper is similar to Ghiassian and Sutton (2021) and considers the same algorithms as theirs but applies the algorithms to two tasks that have 10 times larger state spaces than the Collision task.
The product of importance sampling ratios in the \hp task is larger than the Collision task and the product of the ratios in the \hv task is larger than the \hp task.
We explore the whole parameter space of algorithms and conclude that the problem variance---variance induced by the product of the importance sampling ratios---heavily affects the algorithm performance.

%


\section{Formal Framework}
\label{sct:FormalFramework}

We simulate the agent-environment interaction using the MDP framework.
An agent and an environment interact at discrete time steps, $t=0, 1, \ldots$
At each time step the environment is in a state $S_t\in\S$ and chooses an action, $A_t\in\A$ under a behavior policy $b:~\A \times \S \rightarrow [0, 1]$.
For a state and an action ($s,a$), the probability that action $a$ is taken in state $s$ is denoted by $b(a|s)$ where $``|"$ means that the probability distribution is over $a$ for each $s$.
After choosing an action, the agent receives a numerical reward $R_{t+1}\in\R\subset\Re$ and the environment moves to the next state $S_{t+1}$.
The transition from $S_t$ to $S_{t+1}$ depends on the MDP's transition dynamics.

In off-policy learning, the policy the agent learns about is different from the policy the agent uses for behaviour.
The policy the agent learns about is denoted by $\pi$ and is termed the \emph{target policy}, whereas the policy that is used for behavior is denoted by $b$ and is termed the \emph{behavior policy}.
The goal is to learn the expectation of the sum of the future rewards, the \emph{return}, under a target policy.
Both target and behavior policies are fixed in prediction learning.
The return includes a termination function: $\gamma:~\S\times\A\times\S \rightarrow [0, 1]$:
\begin{align}
G_t & \defeq R_{t+1} + \gamma(S_t, A_t, S_{t+1}) R_{t+2} + \nonumber\\
& \gamma(S_t, A_t, S_{t+1})\gamma(S_{t+1}, A_{t+1}, S_{t+2}) R_{t+3} + \cdots \nonumber 
\end{align}
If for some triplet, $S_k, A_k, S_{k+1}$, the termination function returns zero, the accumulation of the rewards is terminated.

The expectation of the return when starting from a specific state and following a specific policy thereafter, is called the \emph{value} of the state under the policy.
The value function under policy $\pi$ is defined as $v_\pi(s) \defeq \CEpi{G_t}{S_t=s}$,
where $\CEpi{G_t}{S_t}$ is the conditional expectation of the return given that the agent starts in $S_t=s$, and follows $\pi$ thereafter.

In many problems of interest the state space is large and the value function should be approximated using limited resources.
We use linear functions to approximate $v_\pi(s)$.
The approximate value function, $\vhat(\cdot, \vecw)$, is a linear function of a weight vector $\vecw \in \Re^d$.
Corresponding to each state $s$, there is a feature vector in the $d$-dimensional space, $\vecx(s)$, where $d \ll |\S|$.
The approximate value for a state is:
\begin{align}
\vhat(s, \vecw) \defeq \vecw\tr\vecx(s) \approx v_\pi(s). \nonumber \label{eq:ApproximateValueFunction}
\end{align}
%


\section{Algorithms}
\label{sct:Algorithms}

Five Gradient-TD and two Emphatic-TD algorithms, plus Off-policy TD\la, Tree Backup\la (Precup, Sutton, \& Singh, 2000), Vtrace\la (Espeholt et al., 2018), and ABTD\ze (Mahmood, Yu, \& Sutton, 2017) are considered in this study.
From the Gradient-TD family, we consider GTD\la, GTD2\la (Sutton et al., 2009), HTD\la (Maei, 2011; Hackman, 2012; White \& White, 2016), Proximal GTD2\la (Mahadevan et al., 2014; Liu et al., 2015; Liu et al., 2016), and TDRC\la (Ghiassian et al., 2020).
From the Emphatic-TD family we consider Emphatic TD\la (Sutton, Mahmood, \ White, 2016), and Emphatic TD\lab (Hallak et al., 2016).

All algorithms chosen for this study are highly scalable and require linear computation per time step.
All Gradient and Emphatic TD algorithms are guaranteed to converge with linear function approximation.
Tree Backup\la, Retrace\la (control variant of Vtrace\la), and ABQ\ze (control variant of ABTD\ze) all have convergent variants that we chose not to study here.
The convergent variants use gradient correction, need an extra weight vector, and are less practical than the form studied here (Touati et al., 2017).
Off-policy TD\la is not guaranteed to converge but has shown to be effective in previous studies (Geist \& Scherrer, 2014).
For an overview of these algorithms see Ghiassian and Sutton (2021).


\section{\hp Task}
\label{sct:HallwayProximityTask}
\begin{figure}[]
      \centering
      \includegraphics[width=0.7\linewidth]{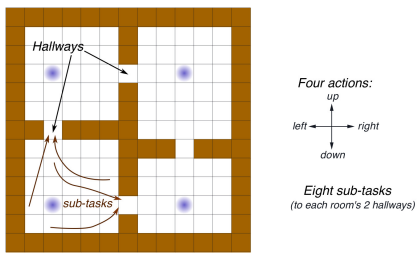}
      \caption{The Four Rooms environment. Four actions are possible in each state. Two hallway states are shown with arrows. Four sub-tasks are also schematically shown. The four shaded states are the ones where the \hp and the \hv tasks have different policies.}
      \label{fig:FourRoomsFigure}
\end{figure}
The \hp task uses a variant of the Four Rooms environment MDP (Sutton, Precup, \& Singh, 1999): a gridworld with 104 states, roughly partitioned into four contiguous areas called rooms (Figure~\ref{fig:FourRoomsFigure}).
These rooms are connected through four hallway states.
Four deterministic actions are available in each state: {\tt left, right, up}, and {\tt down}.
Taking each action results in moving in the corresponding direction except for cells neighboring a wall in which the agent will not move if it takes the action toward the wall.
The task is continuing.

The \hp task consists of eight sub-tasks.
Solving a sub-task corresponds to learning the value function for the target policy corresponding to the sub-task.
Under the corresponding target policy, the agent follows a shortest path to one of the room's hallways, which we refer to as the corresponding hallway.
If two actions are optimal in a state one of the two is taken randomly with equal probability.
For a sample target policy defined in the upper left room, see Figure~\ref{fig:FourRoomsOneRoomPolicy} in Appendix~\ref{app:MoreExperimentalDetails}.
The termination function returns 0.9 while the agent is in the room.
Once the agent reaches the corresponding hallway, the termination function returns zero, without affecting the actual trajectory of the behavior policy.
The termination function remains zero for all states that are not part of the sub-task.
When the agent reaches the corresponding hallway, it receives a reward of +1.
The rewards for all other transitions are zero.
In the same room, another target policy is defined under which the agent follows the shortest path to the other hallway state.
This means that there are exactly two sub-tasks defined for each state, including the hallways.

The \hp task is designed to be a high variance problem.
By a high variance problem, we mean the product of importance sampling ratios can vary between small and large values during learning and can cause large changes in the learned weight vector that might make learning unstable.
Under the target policy, the agent follows a shortest path to a hallway, and the behavior policy is equiprobable random.
If the agent is in the top left state of the upper right room, chooses the {\tt right} action twice, and then chooses the {\tt down} action six times, the product of the importance sampling ratios will become $2^{14}$ because the importance sampling ratio is 2 at the first two time steps and it is $\frac{1}{1/4}=4$ for a total of 6 steps.

The agent learns about two sub-tasks at each time step.
For Emphatic TD, this will be automatically enforced with the interest function set to 0 for all states that are not part of the active sub-task.
Other algorithms do not natively have interest so we enforce this manually by making sure that at each time step, the agent knows what sub-tasks are active and only updates weight vectors corresponding to those sub-tasks.

\begin{figure*}[]
      \centering
      \includegraphics[width=0.7\linewidth]{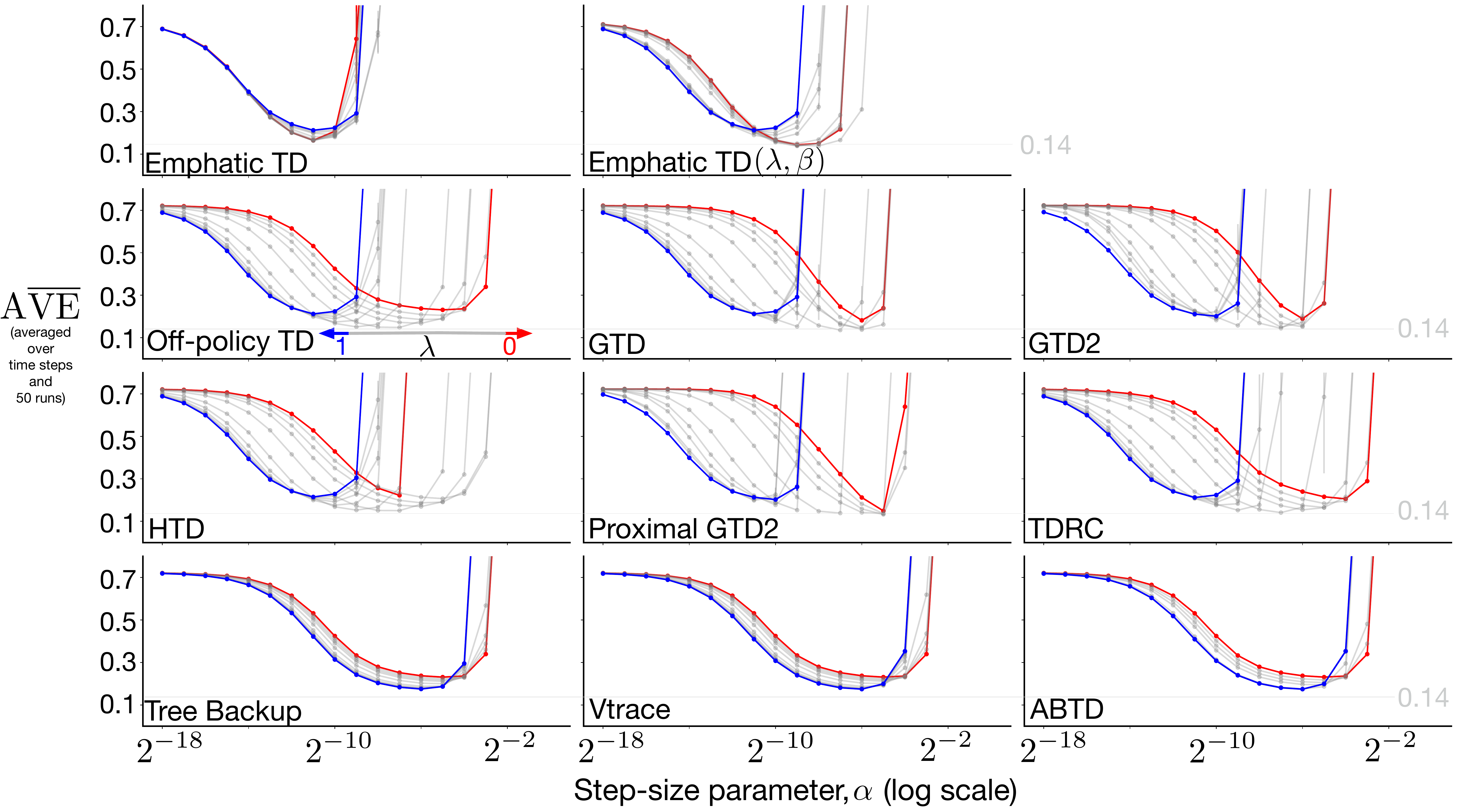}
      \caption{Error as a function of $\alpha$ and \l for all algorithms on the \hp task.
      All algorithms reached the 0.14 error level except Tree Backup\la, Vtrace\la, and ABTD\ze.
      Proximal GTD2\la and Emphatic TD\la were more sensitive to $\alpha$ than other algorithms.
      Emphatic TD\la was less sensitive to \l than other algorithms.}
      \label{fig:MainResultsHP}
\end{figure*}

We used linear function approximation to solve the task.
To represent states, ($x, y$) coordinates were tile coded.
The $x$ coordinate ranged from 0 to 10, where 0 was assigned to the far left cell.
The $y$ coordinate also ranged from 0 to 10, where 0 was the bottom cell.
We used four tilings, each of which was two by two tiles.
In fact, the features used to solve the task can be produced using any system, for example a neural network.
Our focus, in this paper, is on learning the value function linearly using known features, the task that is typically carried out by the last layer of the neural network.

To assess the quality of the value function found by an algorithm, we used the mean squared value error:
\beq
\MSVEm(\w)={\frac{\sum_{s\in\S}\mu_b(s)i(s)\left[\hat{v}(s, \vecw) - v_\pi(s)  \right]^2}{\sum_{s\in\S}\mu_b(s)i(s)}},\nonumber
\eeq
where $i(s)$, the {\em interest function}, $i: \States \rightarrow \{0,1\}$ defines a weighting over states and $\mu_b(s)$ is an approximation of the stationary distribution under the behavior policy which was calculated by having the agent start at the bottom left corner and following the behavior policy for a hundred million time steps and computing the fraction of time the agent spent in each state.
The true value function was calculated by following each of the target policies from each state to their corresponding hallway once.
The interest function is one for all states where the target policy is defined.
Setting $i(s)$ in the error computation ensures that prediction errors from states outside of a room do not contribute to the error computed for each sub-task.
We computed the square root of \MSVE for each policy separately and then simply averaged the errors of the eight approximate value functions to compute an overall measure of error, which we denote by \AVE: $\AVEm(\vecw)\defeq\frac{1}{8}\sum_{j=1}^8 \sqrt{\overline{\text{VE}}(\vecw^{j})}.$


\section{Experimental Setup and Results}
\label{sct:TheHallwayProximityExperiment}

The task and the behavior policy were used to generate 50,000 steps, comprising one \emph{run}.
This was repeated for a total of 50 runs.
The 11 learning algorithms were applied to the 50 runs, each with a range of parameter values.
A list of all parameters used can be found in Table~\ref{tab:params-tbl} in Appendix~\ref{app:ParameterSettings}.
They included 12 values of \l, 19 values of \a, 15 values of $\eta$ (for the Gradient-TD family), six values of $\beta$ (for Emphatic TD\lab), and 12 values of $\zeta$ (for ABTD\ze), for approximately 20,000 algorithm instances in total.
At the beginning of each run, the weight vector was initialized to $\w_0=\vec0$ and then was updated at each step by the algorithm instance.
At each time step, \AVE was computed and recorded.
The code for the experiments can be accessed at: https://github.com/sinaghiassian/OffpolicyAlgorithms.


\paragraph{Main Results}
Performance of an algorithm instance is summarized by one number: \AVE averaged over runs and time steps.
This number is shown for many algorithm instances in Figure~\ref{fig:MainResultsHP}.
Each panel shows one algorithm's performance.

Let us first focus on Off-policy TD\la results shown in the first panel of the second row of Figure~\ref{fig:MainResultsHP}.
This algorithm has two parameters: the step-size parameter, $\alpha$, and the bootstrapping parameter, $\lambda$.
The x-axis shows the value of $\alpha$ at logarithmic scale.
Each curve within the panel shows the performance with one $\lambda$.
The blue curve shows performance with $\lambda=1$ and the red curve shows performance with $\lambda=0$.
Performance with intermediate values of $\lambda$ are shown in gray.
With small $\alpha$, learning was too slow.
With large $\alpha$, divergence happened.
This is why all the curves in the panel are U-shaped.
For each point, the standard error over 50 runs is shown as a bar over the point.
The error bars are often too small and are not visible.

The measure of good performance that we look for in the data is low error over a wide range of parameters.
For Off-policy TD\la, the bottom of the U-shaped curves were at about 0.14 (shown as a thin gray line).
The instances that reached this error level were in a sweet spot.
This sweet spot was large for Off-policy TD\la.

Let us now move on to studying the performance of all algorithms shown in Figure~\ref{fig:MainResultsHP}.
There are lots of similarities between the algorithms.
All algorithms had their best performance with intermediate values of \l, except for Tree Backup\la, Vtrace\la, and ABTD\ze.
All algorithms except the three reached to about 0.14 error level.
With all algorithms, except for Emphatic TD\la, the sweet spot shifted to the left, as \l increased from 0 to 1.
Between the five Gradient-TD algorithms shown in the two middle rows, GTD2 and Proximal GTD2 were more sensitive to $\alpha$ and their U-shaped curves were less smooth than some others.

One of the most distinct behaviors was observed with Emphatic TD\la whose performance changed little as a function of \l.
Its best performance was a bit worse than 0.14 and was achieved with $\lambda=0$.
Emphatic TD\la was more sensitive to $\alpha$ than other algorithms.
Notice how its U-shaped curve is less smooth and narrower at its bottom than many others.

Tree Backup, Vtrace, and ABTD behaved similarly.
They did not reached the 0.14 error level, and had their best performance at \l=1.
Notice how ABTD\ze was less sensitive to its bootstrapping parameter, $\zeta$, and only had three gray curves, whereas Vtrace\la, and Tree Backup\la had more gray curves.
Many of the ABTD\ze gray craves are hidden behind the blue curve.

Overall, on the \hp task, we divide the algorithms into two tiers.
All algorithms except Tree Backup, Vtrace, and ABTD had an error close to 0.14 and are in the first tier.
Tree Backup, Vtrace, and ABTD are in the second tier because regardless of how their parameters were set, they never reached the 0.14 error level.
These conclusions are in some cases similar to the ones from the Collision task (Ghiassian \& Sutton, 2021).
When applied to the Collision task, algorithms were divided into three tiers: Emphatic-TD algorithms were in the top tier, Gradient-TD and Off-policy TD\la were in the middle tier, and Tree Backup, Vtrace, and ABTD were in the bottom tier.
Similar to the Collision task, Tree Backup, Vtrace, and ABTD did not perform as well as other algorithms when applied to the \hp task.
Unlike the Collision task, Emphatic TD's best performance was similar to Gradient-TD algorithms' best performance, but not better.

\begin{figure}[]
      \centering
      \includegraphics[width=\linewidth]{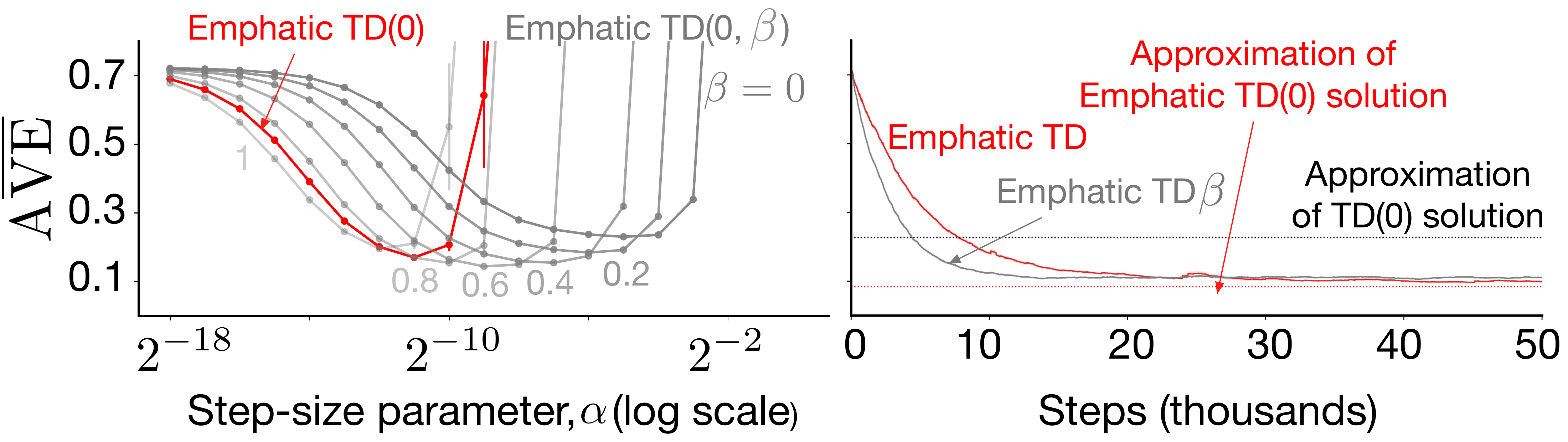}
      \caption{Detail on the Emphatic TD\lab performance on the \hp task at $\lambda=0$ is shown on the left.
      Best learning curves for each algorithm are shown on the right.
      The $\beta$ parameter helped Emphatic TD\lab learn faster.}
      \label{fig:EmphaticsHP}
\end{figure}

\paragraph{Emphatic-TD}
So far, we looked at performance as a function of \a and \l.
We now set $\lambda=0$, and study the effect of $\beta$ on the performance of Emphatic TD\lab.
Errors for various $\beta$s are plotted in the left panel of Figure~\ref{fig:EmphaticsHP}.
Best performance was achieved with an intermediate $\beta$ and was statistically significantly lower than the error of Emphatic TD.

These results make it clear that the $\beta$ parameter improves the performance of Emphatic-TD algorithms on the \hp task.
This is in contrast to the results reported by Ghiassian and Sutton (2021) in which no improvement was observed by varying $\beta$.
It seems like varying $\beta$ might only be useful in cases where the problem variance is high.

Two learning curves and two dashed straight lines are shown in the right panel of Figure~\ref{fig:EmphaticsHP}.
The learning curves correspond to algorithm instances of Emphatic TD(0) and Emphatic TD(0, $\beta$) that minimized the area under the learning curve (AUC).
The dashed lines show the approximate solutions of Emphatic TD(0) and Off-policy TD(0).
These solutions are found using all the data over 50,000 time steps and 50 runs, and the Least-squared algorithms discussed in Appendix~\ref{app:LestSquaredAlgorithmsAsBaselines}.
These solutions show the error level these algorithms would converge to if they were applied to the task with a small enough $\alpha$ and were run for long enough.

Emphatic TD(0) learned slower than Emphatic TD(0, $\beta$).
In fact, Emphatic TD\la learned slower than all other algorithms when applied to the \hp task.
Learning curves for algorithm instances with the smallest AUC for of all algorithms for general \l are shown in Figure~\ref{fig:BestOverallLearningCurves} in Appendix~\ref{app:MoreExperimentalDetails}.
However, if Emphatic TD had enough time to learn, it would converge to a lower asymptotic solution than other algorithms, as shown by the straight dashed lines.

The results from the Collision and \hp task collectively show that Emphatic TD\la tends to have a lower asymptotic error level, but is more prone to the problem variance.
On the Collision task, Emphatic TD\la had a lower asymptotic error level and learned faster than other algorithms.
Moving on to the \hp task, Emphatic TD learned slower than other algorithms, but still had a lower asymptotic error.


\paragraph{Gradient-TD}
To study Gradient-TD algorithms in more detail, we set $\lambda=0$ and analyze the effect of $\eta$ on performance, where $\alpha = \eta * \alpha_\vecv$, and $\alpha_\vecv$ is the second step size.

\begin{figure}[]
      \centering
      \includegraphics[width=0.87\linewidth]{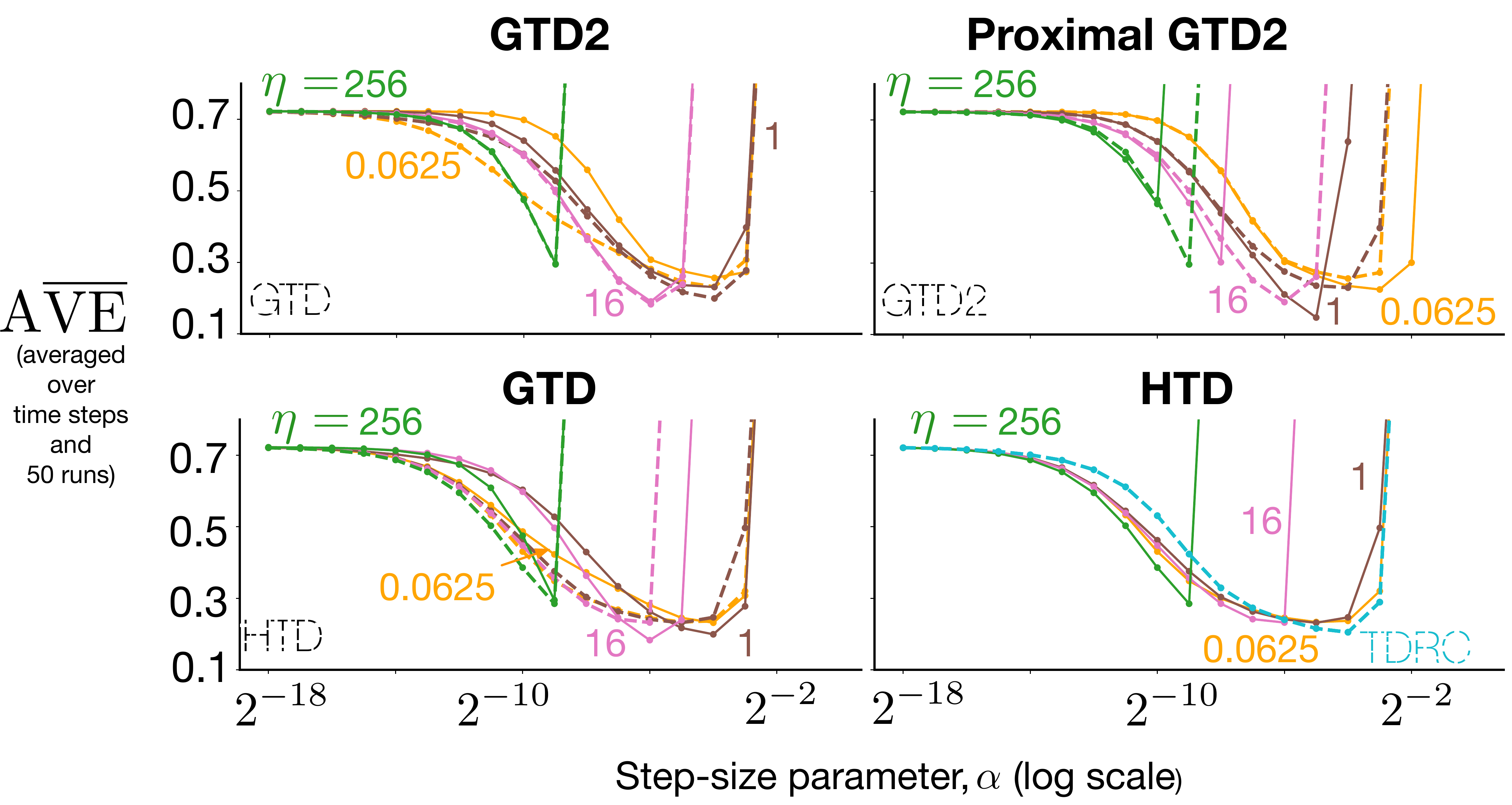}
      \caption{Error as a function of $\alpha$ and $\eta$ at $\lambda=0$ on the \hp task.
      A second algorithm's performance is shown in each panel for comparison.
      Proximal GTD2 had the lowest error but was more sensitive to $\alpha$ than other algorithms. 
      TDRC and HTD had the lowest sensitivity to $\alpha$.}
      \label{fig:GradientsHP}
\end{figure}

Error as a function of $\alpha$ for various values of $\eta$ is shown in Figure~\ref{fig:GradientsHP}.
Each panel shows the performance of an algorithm as a function of $\alpha$ for four values of $\eta$.
The errors of algorithm instances with the same $\eta$ are connected with a line.
Each panel shows results of one algorithm in solid lines.
Each panel additionally shows the performance of one extra algorithm in dashed lines for comparison.
For example, the upper left panel shows the performance of GTD2(0) in solid and GTD(0) in dashed lines.
The dashed lines were always below the solid lines, meaning that GTD(0) had a lower error than GTD2(0) over all parameters.
Difference between GTD(0) and GTD2(0) was largest with small $\eta$.

With a thorough understanding of one panel, let us now move on to comparing the algorithms in all panels.
We first notice that all algorithms solved the problem fairly well.
According to the upper right panel, Proximal GTD2 had the lowest error among all algorithms, but only with one of its parameter settings.
Proximal GTD2 had a lower error than GTD2 for small $\eta$.
For larger $\eta$ the reverse was true.
According to the lower left panel, GTD had a slightly lower error than HTD; however, HTD was less sensitive to $\alpha$, specifically with $\eta=0.0625$.
According to the lower right panel, TDRC's bowl was almost as wide as HTD's widest bowl.
TDRC has one tuned parameter and thus has one curve.

Although Proximal GTD2 performed better than others, it did so with only one parameter setting, and thus the improvement it provides is not of much practical importance.
HTD, GTD, and TDRC all performed well and were robust to the choice of $\alpha$.
TDRC, specifically, with one tuned parameter, is the easiest to use algorithm for solving the \hp task.


%
\begin{figure*}[]
      \centering
      \includegraphics[width=0.7\linewidth]{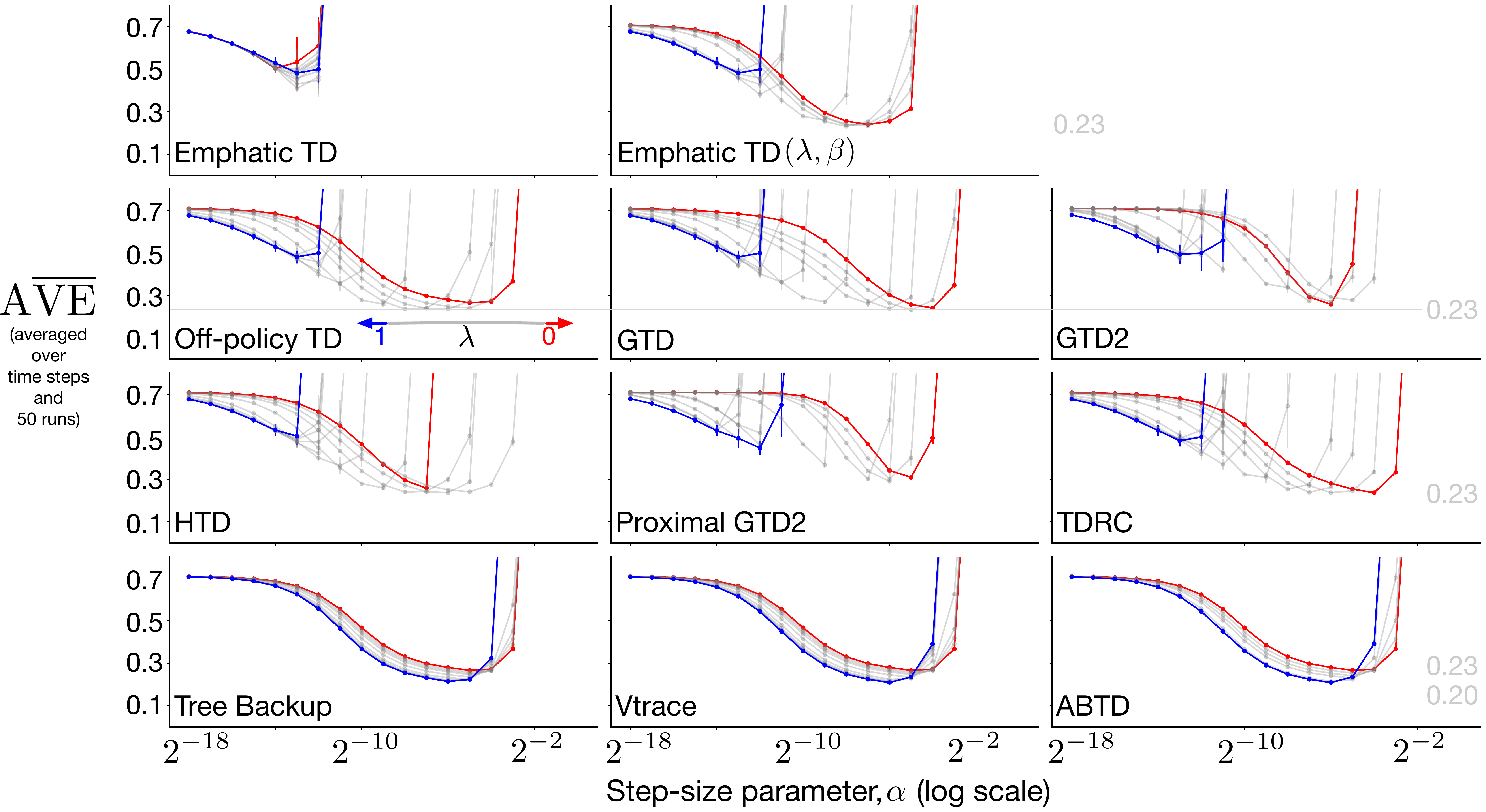}
      \caption{Error as a function of $\alpha$ and \l for all algorithms on the \hv task. 
      Tree Backup\la, Vtrace\la, and ABTD\ze reached the lowest error level (0.2) and were in the top tier.
      All other algorithms except for Emphatic TD\la, Proximal GTD2\la, and GTD2\la reached the 0.23 error-level and were in the middle tier.
      Emphatic TD\la, Proximal GTD2\la, and GTD2\la had a higher error than the rest of the algorithms and were more sensitive to $\alpha$ and were grouped into the bottom tier.}
      \label{fig:MainResultsHVHP}
\end{figure*}

\section{\hv Task}
\label{sct:HighVarianceHallwayTask}

With a slight modification of the \hp task, we increased its variance.
We changed the behavior policy in four states such that one action is chosen with 0.97 and the three other actions with 0.01 probability.
These states are the ones shaded in blue in Figure~\ref{fig:FourRoomsFigure}.
In the two left rooms, the {\tt left} action is chosen with 0.97 probability, and in the two right rooms, the {\tt right} action.
This means that, if the {\tt down} action is chosen in the blue state in the upper right room, the importance sampling will be $\frac{1}{1/100}$.
The new task is called the \hv task.
If the agent starts from the upper left state in the upper right room, takes two {\tt right} actions, and then six {\tt down} actions, the product of importance sampling ratios will be $2^{14}\times 25$.
In addition to more extreme importance sampling ratios, this small change in the behavior policy largely changes the state visitation distribution compared to the \hp task.
The states to the left of the blue states in the two left rooms, and the states to the right of the blue states in the two right rooms are visited more often.
Visitation distributions are shown in Figure~\ref{fig:Dists} in Appendix~\ref{app:MoreExperimentalDetails}.


\section{Experimental Setup and Results}
\label{sct:ExperimentalSetupAndResults}

The experimental setup for this task was the same as the \hp task.
Number of time steps, number of runs, and the algorithm instances applied to the task were all the same.


\paragraph{Main Results}
Main results for the \hv task are plotted in Figure~\ref{fig:MainResultsHVHP}.
The variance caused by the importance sampling ratio impacted all algorithms except Tree Backup\la, Vtrace\la, and ABTD\ze.
These three algorithms reached an error of 0.2 (shown as a thin gray line), which was the lowest error achieved on this task.
Similar to the \hp and the Collision tasks, these three algorithms had their best performance with $\lambda=1$.

Now let us focus on the rest of algorithms in the three first rows of Figure~\ref{fig:MainResultsHVHP}.
All algorithms except Emphatic TD\la, Proximal GTD2\la, and GTD2\la reached the 0.23 error level (shown as a thin gray line).
These three algorithms were sensitive to $\alpha$ and did not perform well.
Emphatic TD\la reached an error of about 0.45 which was significantly higher than the error achieved by any other algorithm.

On this task, we divide the algorithms into three tiers.
The top tier comprises of Tree Backup\la, Vtrace\la, and ABTD\ze whose error was the lowest.
The behavior of these was similar across Collision, \hp, and \hv tasks.
In the middle tier are Off-policy TD\la, GTD\la, HTD\la, and TDRC\la.
These algorithms achieved a slightly higher error than the top tier algorithms but still reasonably solved the task.
The bottom tier comprises of Emphatic TD\la, GTD2\la, and Proximal GTD2\la whose best error level was even higher than second tier algorithms.


\paragraph{Emphatic-TD}
We now turn to studying the Emphatic-TD algorithms in more detail.
We set $\lambda=0$, and study the behavior of Emphatic TD\lab with varying $\beta$.
\begin{figure}[]
      \centering
      \includegraphics[width=\linewidth]{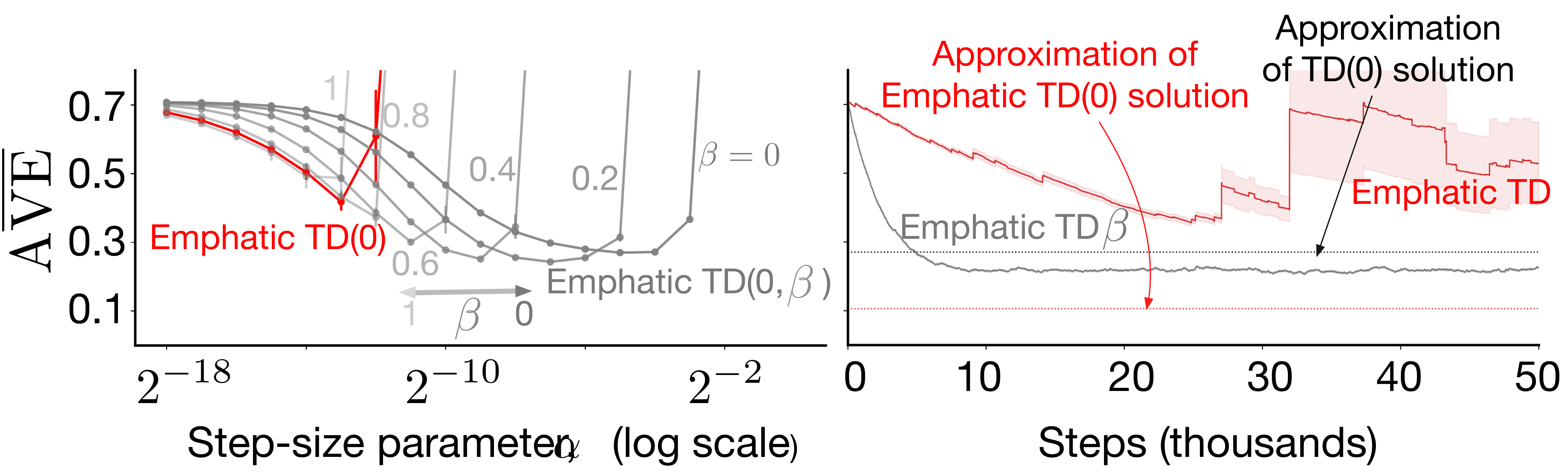}
      \caption{Error as a function of $\alpha$ and $\beta$ at $\lambda=0$ on the \hv task is shown on the left.
      Two best learning curves for Emphatic TD\lab and Emphatic TD\la on the right show that Emphatic TD\lab learned faster.}
      \label{fig:EmphaticsHV}
\end{figure}
Errors for various values of $\beta$ are shown in the left panel of Figure~\ref{fig:EmphaticsHV}.
The best performance was observed with small values of $\beta$.
The bowl was nice and wide with $\beta=0$ and $\beta=0.2$.
After that, with increasing $\beta$, the error consistently increased.

These results show that varying $\beta$ significantly improves Emphatic TD\lab's performance.
Without $\beta$, Emphatic TD(0) performed quite poorly on the \hv task due to large variance.
These results and the results from the \hp task show that $\beta$'s role becomes more salient as the problem variance increases.
On the Collision task, no improvement was observed when varying $\beta$.
On the \hp task, intermediate values of $\beta$ resulted in the best performance, and in the \hv task, small values.
The trend shows that as the problem variance increases, the magnitude of $\beta$ that results in the best performance becomes smaller.

Learning curves for best algorithm instances of Emphatic TD(0) and Emphatic TD(0, $\beta$) are shown in the right panel of Figure~\ref{fig:EmphaticsHV}.
These learning curves correspond to algorithm instances that resulted in minimum AUC.
Emphatic TD(0, $\beta$) learned significantly faster than Emphatic TD.
Emphatic TD(0) did not learn a reasonable approximation of the value function, probably due to being affected by the problem variance.
The two dashed lines in the right panel of the figure show the approximate solutions for Emphatic TD(0) and Off-policy TD(0).
Emphatic TD's solution had a significantly smaller \AVE than Off-policy TD.
This means that if the high variance was not present, Emphatic TD would find a solution with lower error than other algorithms such Off-policy TD\la.

In \hp, \hv , and Collision tasks, Emphatic TD had a lower asymptotic error than other algorithms.
This has also been observed in some previous studies (Ghiassian, Rafiee, \& Sutton, 2016).
However, as the problem variance increases, Emphatic TD\la tends to learn slower.
On the Collision task, it learned the fastest, on the \hp task it learned slower than other algorithms, and it failed to learn a good approximation of the value function in the \hv task.
The trend shows that Emphatic TD has a smaller asymptotic error across tasks but might overall be more prone to the variance issue than other algorithms.


\paragraph{Gradient-TD}
\begin{figure}[]
      \centering
      \includegraphics[width=0.95\linewidth]{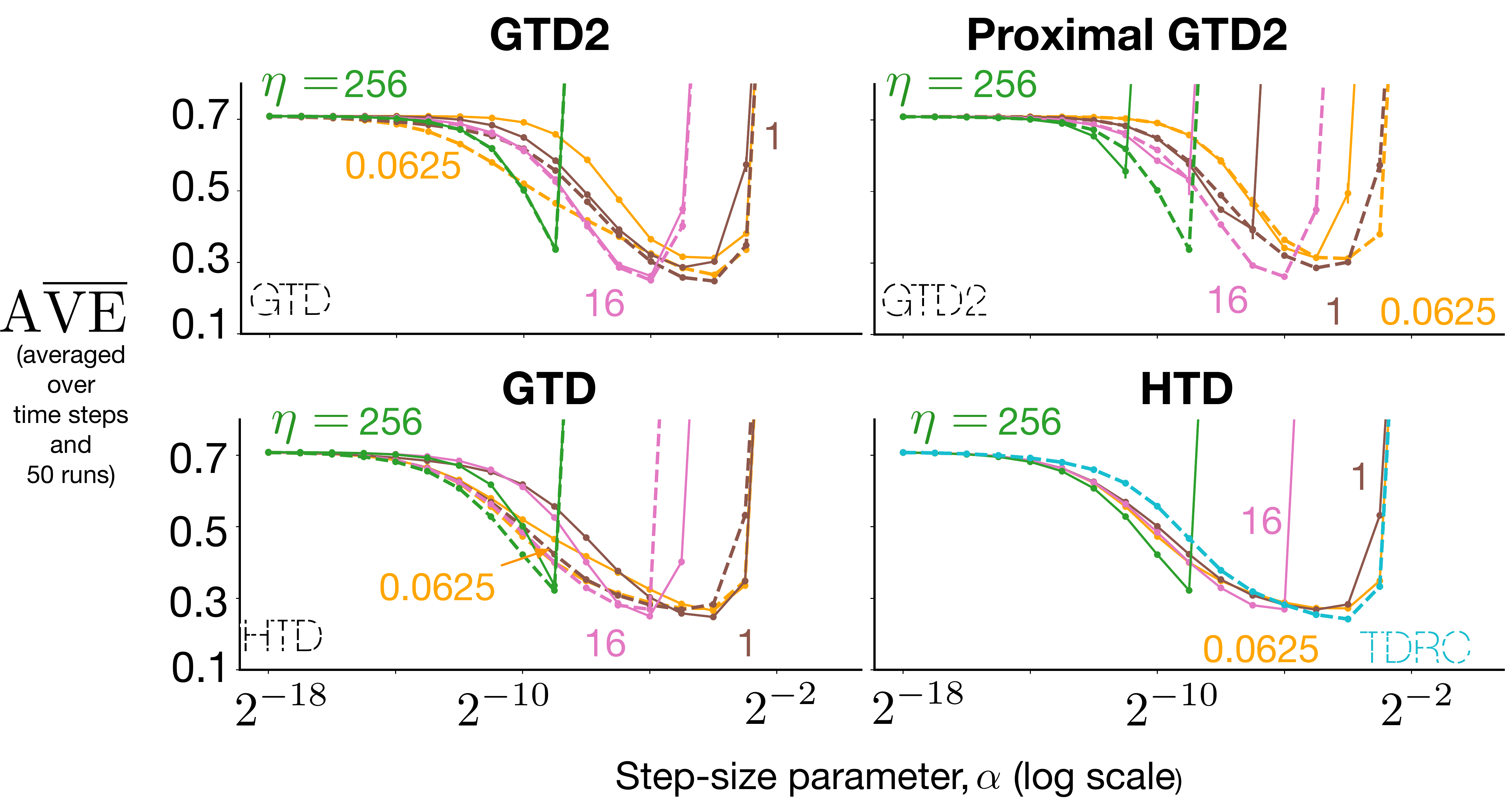}
      \caption{Error as a function of $\alpha$ and $\eta$ at $\lambda=0$ on the \hv task.
      The error of Proximal GTD2 (solid lines in the upper right panel) was higher than others.}
      \label{fig:GradientsHV}
\end{figure}
We now turn to a more detailed analysis of Gradient-TD algorithms on the \hv problem.
Errors for $\lambda=0$ for various $\eta$ are plotted in Figure~\ref{fig:GradientsHV}.

Based on the data provided in the upper left panel, GTD2 generally performed worse than GTD.
Based on the upper right panel, Proximal GTD2 had a significantly larger error than GTD2.
According to the two lower panels, TDRC, HTD, and GTD all performed similarly and were all relatively robust to the choice of $\alpha$.

Let us now summarize the performance of Gradient-TD algorithms across tasks.
On the Collision and \hp tasks, Proximal GTD2 had the lowest error of all Gradient-TD algorithms.
On the \hv task, however, it had a higher error than all Gradient-TD algorithms.
The trend across problems shows that Proximal GTD2 might be able to reach a lower error level than other Gradient-TD algorithms but is more prone to high variance than other Gradient-TD algorithms.
In addition, the lower error level it achieves does not seem to be of much practical utility because it is rare.
GTD\la, HTD\la, and TDRC\la all seem to work well across tasks.
HTD\la seems to be easier to tune across problems (see how its various bowls are smoother and wider than GTD in the lower left panel of Figure~\ref{fig:GradientsHV}).
TDRC seems to be the easiest to use Gradient-TD algorithm because it has one tuned parameter and works as well as HTD across tasks.


\section{Conclusions and Future Directions}
\label{sct:ConclusionsAndFutureDirections}

This study along with the one conducted by Ghiassian and Sutton (2021) paints a detailed picture of algorithms' performance as a function of the problem variance.
Regarding the interplay of the algorithms' performance and variance of the problem, three main points were shown in this paper:
\begin{enumerate*}
\item We showed, for the first time, that Emphatic TD\la tends to have a lower asymptotic error level but it is more prone to the high variance issue than other algorithms.
\item We showed, for the first time, that Proximal GTD2\la seems to be prone to the variance issue as well, but less so than Emphatic TD\la, and
\item We showed, for the first time, that Tree Backup\la, Vtrace\la, and ABTD\ze are most robust to the problem variance but perform worse than other algorithms on simple problems where high variance is not expected.
\end{enumerate*}
Our message for practitioners is to use Tree Backup, Vtrace, or ABTD in problems where high variance is expected, and to use Emphatic TD\la otherwise.

We observed that Emphatic TD\la is the best performing algorithm in terms of asymptotic error, but it is prone to the high variance that might be induced by the product of the importance sampling ratios.
A few approaches might help speed up Emphatic-TD's learning.
One approach might be through meaningfully defining the interest function.
All studies so far in the literature, have used an interest equal to one in all states.
The fact that Emphatic TD is affected by the variance more than other algorithms might be due to the increase in the emphasis variable:
At each time step, the emphasis variable is incremented by the interest of the visited state (see the update rule for Emphatic TD\la in Appendix~\ref{app:PseudoCodes}).
Setting the interest smaller wherever possible will slow down the increase in the emphasis magnitude and might in turn help control the variance and speed up learning.

One undemanding approach to speeding up Emphatic TD might be through step size adaptation.
Other than using optimizers such as Adam (Kingma \& Ba, 2014), sliding step algorithms (Tian, 2019) or procedures for ratcheting down the step size might help.
The schedule for ratcheting down the step size could be determined by an algorithm such as adaptive upper bound heuristic (Dabney \& Barto, 2012).

Off-policy learning has come a long way but still has a long way to go.
Two of the most central challenges of off-policy learning are stability and slow learning.
The stability issue first became evident through Baird's counterexample (Baird, 1995).
Since then, Baird's counterexample has been used numerous times to exhibit various algorithms' stability in practice.
In this paper, we clearly exhibited the variance challenge of off-policy learning.
The tasks introduced in this paper can be used to assess algorithms' capability in learning in a high variance setting.
This paper is limited in that it does not propose an approach for learning fast, but it does make it clear which off-policy prediction learning algorithms face the slow learning and high variance challenge, and to what degree.

\newpage

\section*{Acknowledgements}
This work was funded by DeepMind, the Alberta Machine Intelligence Institute, NSERC, and CIFAR.
We would like to thank Ali Khlili Yegane for help in preparing the source code.
We thank Martha White, Adam White, Kenny Young, and Banafsheh Rafiee for their help and useful feedback.
The computational resources of Compute Canada were essential to conducting this research.

\section*{References}

\begin{list}{}{%
\setlength{\topsep}{0pt}%
\setlength{\leftmargin}{0.2in}%
\setlength{\listparindent}{-0.2in}%
\setlength{\itemindent}{-0.2in}%
\setlength{\parsep}{\parskip}%
}%
{ \small
\item[]
Baird, L. C. (1995). Residual algorithms: Reinforcement learning with function approximation. In \textit{Proceedings of the 12th International Conference on Machine Learning}, pp.~30--37.

\item[]
Boyan, J. A. (1999). Least-squares temporal difference learning. In \textit{Proceedings of the 16th
International Conference on Machine Learning,} pp.~49--56.

\item[]
Bradtke, S. J., Barto, A. G. (1996). Linear least-squares algorithms for temporal difference
learning. \textit{Machine Learning, 22} pp.~33--57.

\item[]
Dabney, W., Barto, A. G. (2012). Adaptive step-size for online temporal difference learning. In \textit{Proceedings of the 26th AAAI Conference on Artificial Intelligence}.
Chicago	

\item[]
Dann, C., Neumann, G., Peters, J. (2014). Policy evaluation with temporal-differences: A survey and comparison. \textit{Journal of Machine Learning Research, 15} pp.~809--883.

\item[]
Espeholt, L., Soyer, H., Munos, R., Simonyan, K., Mnih, V., Ward, T., Doron, Y., Firoiu, V., Harley, T., Dunning, I. and Legg, S. (2018) IMPALA: Scalable distributed Deep-RL with importance weighted actor-learner architectures. In \textit{Proceedings of the 35th International Conference on Machine Learning.} pp.~1407--1416.

\item[]
Geist, M., Scherrer, B. (2014). Off-policy learning with eligibility traces: A survey. \textit{Journal
of Machine Learning Research 15} pp, 289--333.

\item[]
Ghiassian, S., Patterson, A., Garg, S., Gupta, D., White, A., White, M. (2020). Gradient temporal-difference learning with regularized corrections. In \textit{Proceedings of the 37th International Conference on Machine Learning,} pp.~3524--3534.

\item[]
Ghiassian, S., Rafiee, B., Sutton, R. S. (2016). A first empirical study of emphatic temporal difference learning. In \textit{Workshop on Continual Learning and Deep Learning at the Conference on Neural Information Processing Systems.} ArXiv: 1705.04185.

\item[]
Ghiassian, S., Sutton, R. S. (2021). An Empirical Comparison of Off-policy Prediction Learning Algorithms on the Collision Task. ArXiv: 2106.00922.

\item[]
Hackman, L. (2012). \textit{Faster Gradient-TD Algorithms.} MSc thesis, University of Alberta.

\item[]
Hallak, A., Tamar, A., Munos, R., Mannor, S. (2016). Generalized emphatic temporal-difference learning: Bias-variance analysis. In \textit{Proceedings of the 13th AAAI Conference on Artificial Intelligence}, pp.~1631--1637.

\item[]
Hessel, M., Modayil, J., van Hasselt, H., Schaul, T., Ostrovski, G., Dabney, W., Horgan, D., Piot, B., Azar, M., Silver. D. (2018). Rainbow: Combining improvements in deep reinforcement learning. In \textit{Proceedings of the 32nd AAAI conference on artificial intelligence}.

\item[]
Jaderberg, M., Mnih, V., Czarnecki, W. M., Schaul, T., Leibo, J. Z., Silver, D., Kavukcuoglu, K. (2016). Reinforcement learning with unsupervised auxiliary tasks. ArXiv: 1611.05397.
}

\item[]
Jiang, R., Zahavy, T., Xu, Z., White, A., Hessel, M., Blundell, C., van Hasselt, H. (2021). Emphatic Algorithms for Deep Reinforcement Learning. In \textit{Proceedings of the 38th International Conference on Machine Learning.}

\item[]
Kingma, D. P.,  Ba, J. (2014). Adam: A method for stochastic optimization. ArXiv:1412.6980.

\item[]
Liu, B., Liu, J., Ghavamzadeh, M., Mahadevan, S., Petrik, M. (2015). Finite-Sample Analysis of Proximal Gradient TD Algorithms. In \textit{Proceedings of the 31st International Conference on Uncertainty in Artificial Intelligence}, pp. 504--513.

\item[]
Liu, B., Liu, J., Ghavamzadeh, M., Mahadevan, S., Petrik, M. (2016). Proximal Gradient Temporal-Difference Learning Algorithms. In \textit{Proceedings of the 25th International Conference on Artificial Intelligence (IJCAI-16)}, pp. 4195--4199.

\item[]
Mahadevan, S., Liu, B., Thomas, P., Dabney, W., Giguere, S., Jacek, N., Gemp, I., Liu, J. (2014). Proximal reinforcement learning: A new theory of sequential decision making in primal-dual spaces. ArXiv: 1405.6757.

\item[]
Mahmood, A. R., Yu, H., Sutton, R. S. (2017). Multi-step off-policy learning without importance sampling ratios. ArXiv: 1702.03006.

\item[]
Maei, H. R. (2011). \textit{Gradient temporal-difference learning algorithms.} PhD thesis, University of Alberta.

\item[]
Mnih, V., Kavukcuoglu, K., Silver, D., Rusu, A. A., Veness, J., Bellemare, M. G., Graves, A., Riedmiller, M., Fidjeland, A. K., Ostrovski, G., Petersen, S., Beattie, C., Sadik, A., Antonoglou, I., King, H., Kumaran, D., Wierstra, D., Legg, S., Hassabis, D. (2015). Human level control through deep reinforcement learning. \textit{Nature} pp.~529--533.

\item[]
Precup, D., Sutton, R. S., Singh, S. (2000). Eligibility traces for off-policy policy evaluation. In \textit{Proceedings of the 17th International Conference on Machine Learning}, pp.~759--766.

\item[]
Sutton, R. S. (1988). Learning to predict by the algorithms of temporal-differences. \textit{Machine Learning, 3} pp.~9--44.

\item[]
Sutton, R. S., Barto, A. G. (2018). \textit{Reinforcement Learning: An Introduction,} second edition. MIT press.

\item[]
Sutton, R. S., Maei, H. R., Precup, D., Bhatnagar, S., Silver, D., Szepesv\'ari, Cs., Wiewiora, E. (2009). Fast gradient-descent algorithms for temporal-difference learning with linear function approximation. In \textit{Proceedings of the 26th International Conference on Machine Learning}, pp.~993--1000.

\item[]
Sutton, R. S., Mahmood, A. R., White, M. (2016). An emphatic approach to the problem of off-policy temporal- difference learning. \textit{Journal of Machine Learning Research, 17} pp.~1--29.

\item[]
Sutton, R. S., Modayil, J., Delp, M., Degris, T., Pilarski, P. M., White, A., Precup, D. (2011). Horde: A scalable real-time architecture for learning knowledge from unsupervised sensorimotor interaction. In \textit{Proceedings of the 10th International Conference on Autonomous Agents and Multiagent Systems}, pp.~761--768.

\item[]
Sutton, R. S., Precup, D., Singh, S. (1999). Between MDPs and semi-MDPs: A framework for temporal abstraction in reinforcement learning. \textit{Artificial Intelligence, 112} pp.~181--211.

\item[]
Thomas, P. S. (2015). \textit{Safe reinforcement learning.} PhD thesis, University of Massachusetts Amherst.

\item[]
Tian, T. (2018). \textit{Extending the Sliding-step Technique of Stochastic Gradient Descent to Temporal Difference Learning}. PhD thesis, University of Alberta.

\item[]
Touati, A., Bacon, P. L., Precup, D., Vincent, P. (2017). Convergent tree-backup and retrace with function approximation. ArXiv: 1705.09322.

\item[]
van Hasselt, H., Guez, A., Silver, D. (2016). Deep reinforcement learning with double Q-learning. In \textit{proceedings of the 30th AAAI conference on artificial intelligence}.

\item[]
Watkins, C. J. C. H. (1989). \textit{Learning from delayed rewards.} PhD thesis, University of Cambridge.

\item[]
White, A., White, M. (2016). Investigating practical linear temporal difference learning. In
\textit{Proceedings of the 2016 International Conference on Autonomous Agents and Multiagent Systems,} pp.~494--502.

\item[]
Young, K., Tian, T. (2019). Minatar: An atari-inspired testbed for thorough and reproducible reinforcement learning experiments. ArXiv:1903.03176.
\end{list}

\appendix

\clearpage

\section{Update Rules}
\label{app:PseudoCodes}

In this section, update rules for all algorithms studied in this paper are provided as a single point of reference.
\newline

\noindent\textbf{TD\la:}
\begin{align*}
\delta_t \defeq&\,\, R_{t+1} + \gamma_{t+1} \vecw_{t}^{\tr}\vecx_{t+1} - \vecw_{t}^{\tr}\vecx_{t}\\
\vecz_t \leftarrow& ~\rho_t(\gamma_t \lambda_t \vecz_{t-1}  +\vecx_t) \textnormal{\qquad with } \vecz_{-1} = \bf{0}\nonumber\\
\vecw_{t+1} \leftarrow& ~ \vecw_t +\alpha \delta_t \vecz_t
\end{align*}
\newline
\noindent\textbf{GTD\la:}
\begin{align*}
\delta_t \defeq&\,\, R_{t+1} + \gamma_{t+1}\vecw_t^{\tr}\vecx_{t+1} - \vecw_t^{\tr}\vecx_{t}\nonumber\\
\vecz_t \leftarrow& ~\rho_t(\gamma_t \lambda_t \vecz_{t-1}  +\vecx_t) \textnormal{\qquad with } \vecz_{-1} = \bf{0}\nonumber\\
\vecv_{t+1} \leftarrow& ~\vecv_t + \alpha_\vecv\biggl[\delta_t\vecz_t - (\vecv_t^{\tr}\vecx_t)\vecx_{t}  \biggr] \nonumber\\\
\vecw_{t+1} \leftarrow& ~\vecw_t + \alpha \delta_t \vecz_t - \alpha \gamma_{t+1} (1-\lambda_{t+1})(\vecv_t^{\tr}\vecz_t)\vecx_{t+1}
\end{align*}

\noindent\textbf{TDRC\la:}
\begin{align*}
\delta_t \defeq&\,\, R_{t+1} + \gamma_{t+1}\vecw_t^{\tr}\vecx_{t+1} - \vecw_t^{\tr}\vecx_{t}\nonumber\\
\vecz_t \leftarrow& ~\rho_t(\gamma_t \lambda_t \vecz_{t-1}  +\vecx_t) \textnormal{\qquad with } \vecz_{-1} = \bf{0}\nonumber\\
\vecv_{t+1} \leftarrow& ~\vecv_t + \alpha\biggl[\delta_t\vecz_t - (\vecv_t^{\tr}\vecx_t)\vecx_{t}  \biggr] - \alpha\vecv_t \nonumber\\\
\vecw_{t+1} \leftarrow& ~\vecw_t + \alpha \delta_t \vecz_t - \alpha \gamma_{t+1} (1-\lambda_{t+1})(\vecv_t^{\tr}\vecz_t)\vecx_{t+1}
\end{align*}

\noindent\textbf{GTD2\la:}
\begin{align*}
\delta_t \defeq&\,\, R_{t+1} + \gamma_{t+1}\vecw_t^{\tr}\vecx_{t+1} - \vecw_t^{\tr}\vecx_{t}\nonumber\\
\vecz_t \leftarrow& ~\rho_t(\gamma_t \lambda_t \vecz_{t-1}  +\vecx_t) \textnormal{\qquad with } \vecz_{-1} = \bf{0}\nonumber\\
\vecv_{t+1} \leftarrow& ~\vecv_t + \alpha_\vecv\biggl[\delta_t\vecz_t - (\vecv_t^{\tr}\vecx_t)\vecx_{t}  \biggr] \nonumber\\
\vecw_{t+1} \leftarrow& ~\vecw_t +\alpha (\vecv_t^{\tr}\vecx_t)\vecx_t - \alpha \gamma_{t+1} (1 - \lambda_{t+1}) (\vecv_t^{\tr}\vecz_t)\vecx_{t+1}
\end{align*}

\noindent\textbf{HTD\la:}
\begin{align*}
\delta_t \defeq&\,\, R_{t+1} + \gamma_{t+1}\vecw_t^{\tr}\vecx_{t+1} - \vecw_t^{\tr}\vecx_{t}\nonumber\\
\vecz_t \leftarrow& ~\rho_t(\gamma \lambda \vecz_{t-1}^{\rho}  +\vecx_t) \textnormal{\qquad with } \vecz_{-1} = \bf{0}\nonumber\\
\vecz_t^b \leftarrow& ~\gamma \lambda \vecz_{t-1}  +\vecx_t  \textnormal{\qquad with } \vecz_{-1}^b = \bf{0} \nonumber\\
\vecv_{t+1} \leftarrow& ~\vecv_t + \alpha_\vecv\biggl[\delta_t\vecz_t - (\vecx_t - \gamma_{t+1}\vecx_{t+1})(\vecv_t^{\tr}\vecz_t^b)  \biggr] \nonumber\\\
\vecw_{t+1} \leftarrow& ~\vecw_t + \alpha\biggl[\delta_t\vecz_t + (\vecx_t - \gamma_{t+1}\vecx_{t+1})(\vecz_t -\vecz_t^b)^{\tr}\vecv_t  \biggr]
\end{align*}

\noindent\textbf{Proximal GTD2\la:}
\begin{align*}
\delta_t \defeq&\,\, R_{t+1} + \gamma_{t+1}\vecw_t^{\tr}\vecx_{t+1} - \vecw_t^{\tr}\vecx_{t}\nonumber\\
\vecz_t \leftarrow& ~\rho_t(\gamma_t \lambda_t \vecz_{t-1}  +\vecx_t)\textnormal{\quad with } \vecz_{-1} = \bf{0}\nonumber\\
\vecv_{t+\frac{1}{2}} \leftarrow& ~\vecv_{t} + \alpha_\vecv \biggl[\delta_{t} \vecz_t - (\vecv_{t}^{\tr}\vecx_t)\vecx_t\biggr]\\
\vecw_{t+\frac{1}{2}} \leftarrow& ~\vecw_t +\alpha (\vecv_t^{\tr}\vecx_t)\vecx_t -\alpha \gamma_{t+1} (1 - \lambda_{t+1}) (\vecv_t^{\tr}\vecz_{t}^{\rho} ) \vecx_{t+1}\\
\delta_{t+\frac{1}{2}} \defeq&\,\, R_{t+1} + \gamma_{t+1} \vecw_{t+\frac{1}{2}}^{\tr}\vecx_{t+1} - \vecw_{t+\frac{1}{2}}^{\tr}\vecx_{t}\\
\vecv_{t+1} \leftarrow& ~\vecv_{t} + \alpha_\vecv \biggl[\delta_{t+\frac{1}{2}} \vecz_t^\rho - (\vecv_{t+\frac{1}{2}}^{\tr}\vecx_t)\vecx_t\biggr]\\
\vecw_{t+1} \leftarrow& ~\vecw_t +\alpha (\vecv_{t+\frac{1}{2}}^{\tr}\vecx_t)\vecx_t -\alpha  \gamma_{t+1} (1 - \lambda_{t+1}) (\vecv_{t+\frac{1}{2}}^{\tr}\vecz_{t}^{\rho} ) \vecx_{t+1}
\end{align*}
\newline

\noindent\textbf{Emphatic TD\la:}
\begin{align*}
\delta_t \defeq&\,\, R_{t+1} + \gamma_{t+1}\vecw_t^{\tr}\vecx_{t+1} - \vecw_t^{\tr}\vecx_{t}\nonumber\\
F_t \leftarrow& ~\rho_{t-1}\gamma_t F_{t-1} + I_t \textnormal{\quad with } F_{0} = I_0\\
M_{t}  ~\defeq&\,\, \lambda_t I_t + (1 - \lambda_t)F_t\\
\vecz_t \leftarrow& ~\rho_t \left(\gamma_t \lambda \vecz_{t-1} + M_t \vecx_{t}\right)\textnormal{\quad with } \vecz_{-1} = \bf{0}\\
\vecw_{t+1} \leftarrow& ~ \vecw_t +\alpha \delta_t \vecz_t
\end{align*}
\newline

\noindent\textbf{Emphatic TD($\lambda\, ,\beta$):}
\begin{align*}
\delta_t \defeq&\,\, R_{t+1} + \gamma_{t+1}\vecw_t^{\tr}\vecx_{t+1} - \vecw_t^{\tr}\vecx_{t}\nonumber\\
F_t \leftarrow& ~\rho_{t-1}\beta F_{t-1} + I_t \textnormal{\quad with } F_{0} = I_0\\
M_{t}  \defeq&\,\, \lambda_t I_t + (1 - \lambda_t)F_t\\
\vecz_t \leftarrow& ~\rho_t \left(\gamma_t \lambda \vecz_{t-1} + M_t \vecx_{t}\right)\textnormal{\quad with } \vecz_{-1} = \bf{0}\\
\vecw_{t+1} \leftarrow& ~ \vecw_t +\alpha \delta_t \vecz_t
\end{align*}

\noindent\textbf{Tree Backup\la for prediction:}
\begin{align*}
\delta_t^{\rho} \defeq&\,\, \rho_t \biggl( R_{t+1} + \gamma_{t+1}\vecw_t^{\tr}\vecx_{t+1} - \vecw_t^{\tr}\vecx_{t}\biggr)\nonumber\\
\vecz_t \leftarrow& ~\gamma_t \lambda_t \pi_{t-1} \vecz_{t-1} + \vecx_{t}\textnormal{\quad with } \vecz_{-1} = \bf{0}\\
\vecw_{t+1} \leftarrow& ~\vecw_t +\alpha \delta_t^{\rho} \vecz_t
\end{align*}

\noindent\textbf{Vtrace\la:}
\begin{align*}
\delta_t \defeq&\,\, R_{t+1} + \gamma_{t+1}\vecw_t^{\tr}\vecx_{t+1} - \vecw_t^{\tr}\vecx_{t}\nonumber\\
\vecz_t \leftarrow& ~\textnormal{max}(\rho_t, 1) \left(\gamma_t \lambda \vecz_{t-1} + \vecx_{t}\right)\textnormal{\quad with } \vecz_{-1} = \bf{0}\\
\vecw_{t+1} \leftarrow& ~\vecw_t +\alpha \delta_t \vecz_t
\end{align*}

\noindent\textbf{ABTD($\zeta$):}
\begin{align*}
\delta_t^{\rho} \defeq&\,\, \rho_t \biggl( R_{t+1} + \gamma_{t+1}\vecw_t^{\tr}\vecx_{t+1} - \vecw_t^{\tr}\vecx_{t}\biggr)\nonumber\\
\vecz_t \leftarrow& ~\gamma_t \nu_{t-1} \pi_{t-1} \vecz_{t-1} + \vecx_{t}\textnormal{\quad with } \vecz_{-1} = \bf{0}\\
\vecw_{t+1} \leftarrow& ~ \vecw_t +\alpha \delta_t^{\rho} \vecz_t
\end{align*}

\section{Least-squared Algorithms as Baselines}
\label{app:LestSquaredAlgorithmsAsBaselines}
Temporal-difference learning algorithms, when they converge, satisfy the Bellman equation (Sutton \& Barto, 2018); a fixed point solution that can be directly computed with least-squared algorithms (Bradtke \& Barto, 1996; Boyan, 1999).
In fact, TD\la, Gradient-TD algorithms, and all algorithms used in this paper except Emphatic TD\la, Emphatic TD\lab, Tree Backup\la, Vtrace\la, and ABTD\ze converge to the minimum of the mean squared projected Bellman error ($\PBE$), also known as the TD\la fixed point. 
In the case of a fixed basis, we can analytically solve the $\PBE$ for the weights that satisfy the fixed point equation.
The algorithm that solves the least-squared problem to find the fixed point of TD\la is called least Squared Temporal-Difference or LSTD\la.
The weight vector computed by LSTD\la (from a finite batch of training data) represents the weight vector that TD\la would converge to with repeated presentation of online data.

Different weightings of the $\PBE$ can be simply incorporated into LSTD\la. 
For example, combining LSTD\la and the emphatic weighting produces the weight vector that Emphatic TD\la would converge to, given a fixed batch of data.
Although, LSTD\la can be updated online and incrementally, its complexity is different from the algorithms that we used in our empirical study in this paper.
Least-squared algorithms are only of interest as baselines in this paper because their computation is quadratic in the number of weights ($O(n^2)$) and are in general problematic with large state representations (like for example representations that might be learned using a Neural Network), compared to the algorithms that are presented so far in the paper such as Gradient-TD methods that require linear computation per time-step in the number of weights ($O(n)$).

To compute the LSTD\la fixed-point, we first find the value of $\vecw$ for which the fixed-point TD\la is zero; $\vecw = \Amat^{-1}\vecb$.
The values of the $\Amat$ matrix and the $\vecb$ vector can be estimated incrementally using the following update rules:
\begin{align}
\vecz_t &\leftarrow \rho_t\gamma_t(\lambda_t \vecz_{t-1} + \vecx_t), \nonumber\\
\Amat_{t+1} &\leftarrow \Amat_t + \frac{1}{t+1}[\vecz_t(\vecx_t - \gamma_{t+1}\vecx_{t+1})\tr - \Amat_t], \nonumber\\
\vecb_{t+1} &\leftarrow \vecb_t + \frac{1}{t+1}[R_{t+1}\vecz_t - \vecb_t] \nonumber.
\end{align}
\noindent where $t$ is the time starting from 0, $\vecz$ is the eligibility trace and $\rho$ is the importance sampling ratio.

For each experiment, we used the stream of experience (50 runs and 50,000 time steps) we previously generated, to estimate the values of $\Amat$ and $\vecb$.
We then compute the weight vector $\vecw$.
We can similarly find the LSTD\la fixed-point with emphatic weighting by simply changing the trace update rule mentioned above, to:
\begin{align}
F_t &\leftarrow \beta \rho_{t-1} F_{t-1} + I_t, \nonumber\\
M_t &\leftarrow \lambda_t  I_t + (1 - \lambda_t) F_t, \nonumber\\
\vecz_t &\leftarrow \rho_t(\gamma_t\lambda_t \vecz_{t-1} + \vecx_t M_t) \nonumber.
\end{align}
\noindent where $F$ is the \emph{follow-on trace}, and $M$ is the \emph{emphasis}.


\section{More Experimental Details}
\label{app:MoreExperimentalDetails}

\begin{figure}[h!]
      \centering
      \includegraphics[width=0.5\linewidth]{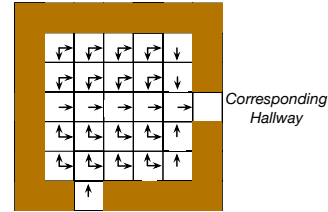}
      \caption{Target policy of the upper left room leading to one of the hallways.}
      \label{fig:FourRoomsOneRoomPolicy}
\end{figure}

\paragraph{Target policy for the upper left room of the \hp task}
The target policy for the \hp task, for the upper left room leading to one of the hallways is shown in Figure~\ref{fig:FourRoomsOneRoomPolicy}.
In some states, there is only one action that takes the agent closer to the hallway.
In those states, the agent takes that action with probability 1.0.
In some states, there are two actions that take the agent one step closer to the hallway.
In those states, the agent takes one of the two actions with equal probability.

\begin{figure}[h!]
      \centering
      \includegraphics[width=\linewidth]{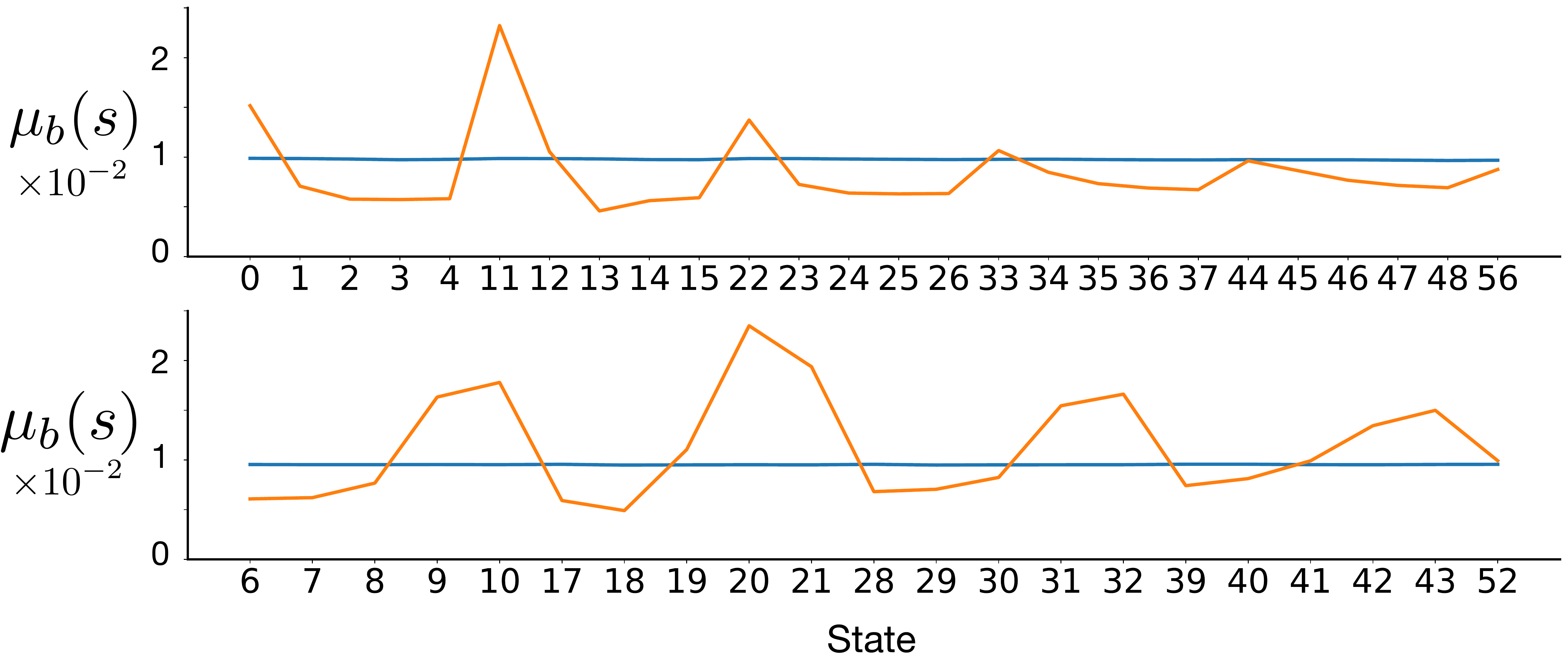}
      \caption{State visitation distribution for \hp and \hv tasks for two of the sub-tasks that are active in the two lower rooms shown in blue and orange respectively.
      The numbers on the x-axis are the state numbers with 0 being the bottom left state, and the state immediately to the right of state 0 being state 1.
      The state immediately above state 0 is state 11.}
      \label{fig:Dists}
\end{figure}

\paragraph{Visitation distribution in \hp and \hv tasks}
The changes we made in the behavior policy in the \hp task to create the \hv task results in changes in the state visitation distribution induced by the behavior policy.
The visitation distribution for two of the sub-tasks that are active in the lower left and lower right rooms are shown in Figure~\ref{fig:Dists}.
As expected, in the \hp task, the visitation distribution was almost uniform and was a bit lower than 0.01 for all states. This makes sense because there are in total 104 states, all which are visited nearly equally.
For the \hv task, the situation is different.
The visitation distribution is higher for some states and lower for some others.
See Figure~\ref{fig:Dists}.
The states are numbered from left to right and bottom to top, meaning that, for example, the lower left state will be state 0, and the leftmost state in the second row from the bottom will be state 11.
For the sub-task active in the lower left room, the visitation distribution was highest for state 11.
This makes sense because the behavior policy chose the left action 97\% of the time it was in state 12.
This means the visitation distribution should be lower for some other states such as state 13, so that the sum of the visitation probabilities becomes 1.

\begin{figure}[h!]
      \centering
      \includegraphics[width=\linewidth]{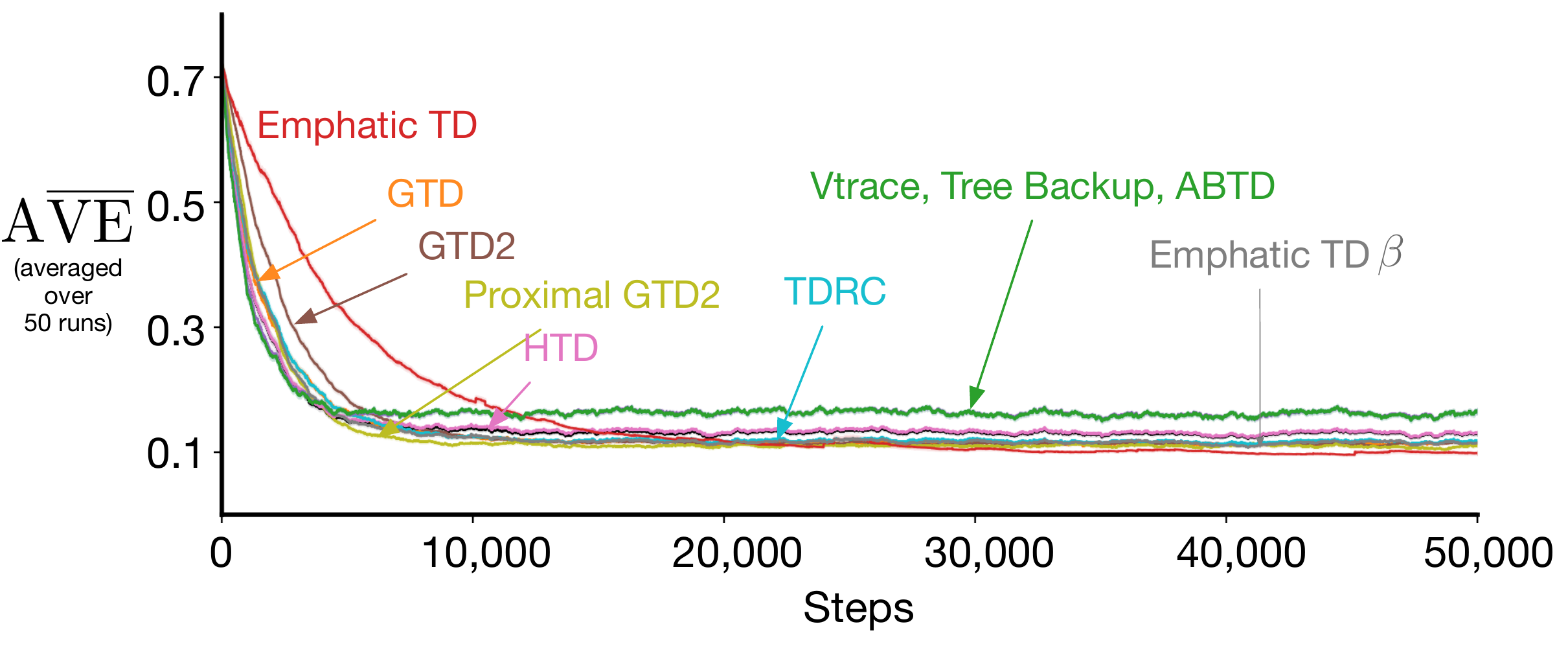}
      \caption{Learning curves for the best algorithm instances of each learning algorithm in the \hp task.}
      \label{fig:BestOverallLearningCurves}
\end{figure}

\paragraph{Learning curves for the best instances of all algorithms in the \hp task}
Learning curves for the best algorithm instances of all learning algorithms are shown in Figure~\ref{fig:BestOverallLearningCurves}.
The best algorithm instances are the ones that had the smallest area under the learning curve.
All parameters including the step-size parameter, bootstrapping parameter, second step-size parameter for Gradient-TD algorithms, and $\beta$ of Emphatic TD\lab were selected such that the area under the learning curve is minimized.
Emphatic TD\la learned significantly slower than the rest of the algorithms on this task.

\begin{figure}[h!]
      \centering
      \includegraphics[width=\linewidth]{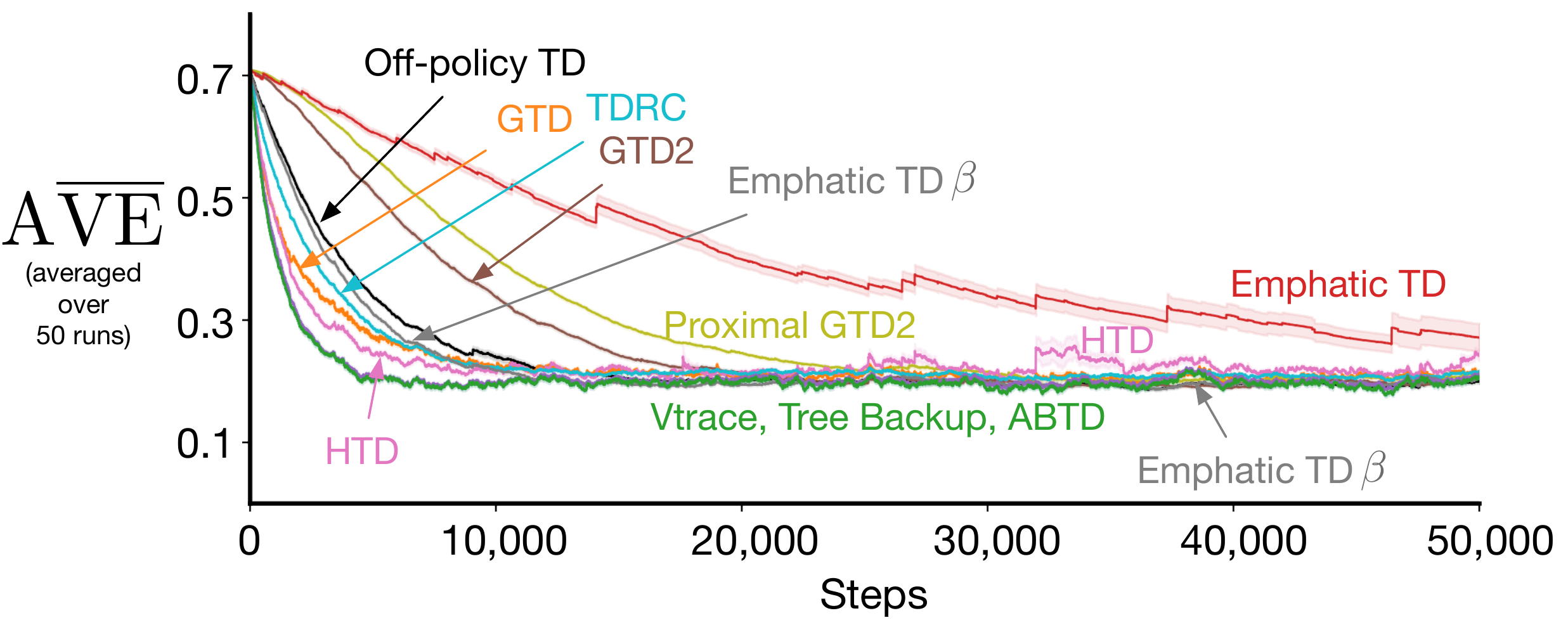}
      \caption{Best algorithm instances of each learning algorithm on the \hv task.}
      \label{fig:BestOverallLearningCurvesHV}
\end{figure}

\paragraph{Learning curves for the best instances of all algorithms in the \hv task}
Learning curves for the best algorithm instances of all learning algorithms are shown in Figure~\ref{fig:BestOverallLearningCurvesHV}.
Emphatic TD\la learned significantly slower than the rest of the algorithms, followed by Proximal GTD2\la and GTD2\la.
The fastest learning algorithms on this task were Tree Backup\la, Vtrace\la, and ABTD\ze.

\paragraph{Specification of Dependencies and the Source Code}

For all experiments conducted in this study, a supercomputer with 2,024 nodes was used (unless mentioned otherwise).
Each node consisted of 2 sockets with 20 intel Skylake cores (2.4 GHz, AVX512), for a total of 40 cores per node.
Each node had 202 GB of RAM.
All experiments were conducted in the Linux CentOS 7 operating system.

Python 3.6, and numpy 1.19.0 was used to run the code.
Matplotlib version 3.2.2 was used to plot the results.

For more details, see the code attached to the submission as a zip file.

\section{Parameter Settings}
\label{app:ParameterSettings}

A list of all algorithm instances (all parameters used in this study) are listed in Table~\ref{tab:params-tbl}.

\begin{table*}[]
\centering
\caption{List of all parameters used in the experiments.}
\begin{tabular}{|c|c|c|c|c|}
\hline
\multicolumn{2}{|c|}{\textbf{\begin{tabular}[c]{@{}c@{}}$~$\\ Algorithms\\ $~$\end{tabular}}} & \textbf{$\eta$ or $\beta$} & \textbf{$\lambda$ or $\zeta$} & \textbf{$\alpha$} \\ \hline
\multicolumn{2}{|c|}{\begin{tabular}[c]{@{}c@{}}$~$\\ Off-policy TD($\lambda$)\\ $~$\end{tabular}} & --- & \multirow{11}{*}{\begin{tabular}[c]{@{}c@{}}0, 0.1, 0.2, \\ 0.3, 0.5, 0.9, 1\\ and\\ 1 - $2^{-x}$\\ where\\ $x \in \{$\\ $2, 3, 4, 5, 6\}$\end{tabular}} & \multirow{11}{*}{\begin{tabular}[c]{@{}c@{}}$\alpha = 2^{-x}$\\ where\\ $x \in \{$\\ $0, 1, 2, \cdots, $\\ $17, 18\}$\end{tabular}} \\ \cline{1-3}
\multirow{5}{*}{\begin{tabular}[c]{@{}c@{}}Gradient-TD\\ Algorithms\end{tabular}} & \begin{tabular}[c]{@{}c@{}}$~$\\ GTD($\lambda$)\\ $~$\end{tabular} & \multirow{4}{*}{\begin{tabular}[c]{@{}c@{}}$2^{x}$\\ where\\ $x \in \{-6, -5, \cdots, 7, 8 \}$\end{tabular}} &  &  \\ \cline{2-2}
 & \begin{tabular}[c]{@{}c@{}}$~$\\ GTD2($\lambda$)\\ $~$\end{tabular} &  &  &  \\ \cline{2-2}
 & \begin{tabular}[c]{@{}c@{}}$~$\\ HTD($\lambda$)\\ $~$\end{tabular} &  &  &  \\ \cline{2-2}
 & \begin{tabular}[c]{@{}c@{}}$~$\\ Proximal GTD2($\lambda$)\\ $~$\end{tabular} &  &  &  \\ \cline{2-3}
 & \begin{tabular}[c]{@{}c@{}}$~$\\ TDRC($\lambda$)\\ $~$\end{tabular} & --- &  &  \\ \cline{1-3}
\multirow{2}{*}{\begin{tabular}[c]{@{}c@{}}Emphatic-TD\\ Algorithms\end{tabular}} & \begin{tabular}[c]{@{}c@{}}$~$\\ Emphatic\\ TD($\lambda$)\\ $~$\end{tabular} & --- &  &  \\ \cline{2-3}
 & \begin{tabular}[c]{@{}c@{}}$~$\\ Emphatic\\ TD($\lambda,~\beta$)\\ $~$\end{tabular} & \begin{tabular}[c]{@{}c@{}}$\beta\in$\\ \{0.0, 0.2, 0.4, 0.6, 0.8, 1.0\}\end{tabular} &  &  \\ \cline{1-3}
\multirow{3}{*}{\begin{tabular}[c]{@{}c@{}}Variable-$\lambda$\\ Algorithms\end{tabular}} & \begin{tabular}[c]{@{}c@{}}$~$\\ Tree Backup($\lambda$)\\ $~$\end{tabular} & \multirow{3}{*}{---} &  &  \\ \cline{2-2}
 & \begin{tabular}[c]{@{}c@{}}$~$\\ Vtrace($\lambda$)\\ $~$\end{tabular} &  &  &  \\ \cline{2-2}
 & \begin{tabular}[c]{@{}c@{}}$~$\\ ABTD($\zeta$)\\ $~$\end{tabular} &  &  &  \\ \hline
\end{tabular}
\label{tab:params-tbl}
\end{table*}

\end{document}